\documentclass[10pt,twocolumn,letterpaper]{article}

\usepackage{cvpr}              

\usepackage{algorithm}
\usepackage{algpseudocode}
\usepackage[dvipsnames]{xcolor}
\usepackage{multirow}
\usepackage{makecell}
\usepackage{svg}
\usepackage{mathbbol}
\usepackage[T1]{fontenc}

\usepackage{tikz}
\usetikzlibrary{positioning}
\usetikzlibrary{shapes.geometric}
\usetikzlibrary{calc}
\def\BibTeX{{\rm B\kern-.05em{\sc i\kern-.025em b}\kern-.08em
    T\kern-.1667em\lower.7ex\hbox{E}\kern-.125emX}}

\usepackage[dvipsnames]{xcolor}

\definecolor{cvprblue}{rgb}{0.21,0.49,0.74}
\usepackage[pagebackref,breaklinks,colorlinks,citecolor=cvprblue]{hyperref}

\def\paperID{39} 
\def\confName{CVPR}
\def\confYear{2024}

\title{Continual Learning with Weight Interpolation}

\author{J\k{e}drzej Kozal\\
Wrocław University of Science and Technology \\
Wrocław, Poland\\
{\tt\small jedrzej.kozal@pwr.edu.pl}
\\
Bartosz Krawczyk\\
Rochester Institute of Technology \\
Rochester NY, USA\\
{\tt\small bartosz.krawczyk@rit.edu}
\and
Jan Wasilewski\\
Rochester Institute of Technology \\
Rochester NY, USA\\
{\tt\small jw7630@g.rit.edu}
\\
Michał Woźniak\\
Wrocław University of Science and Technology\\
Wrocław, Poland\\
{\tt\small michal.wozniak@pwr.edu.pl}
}

\begin{document}
\maketitle

\begin{abstract}
Continual learning poses a fundamental challenge for modern machine learning systems, requiring models to adapt to new tasks while retaining knowledge from previous ones. Addressing this challenge necessitates the development of efficient algorithms capable of learning from data streams and accumulating knowledge over time. This paper proposes a novel approach to continual learning utilizing the weight consolidation method. Our method, a simple yet powerful technique, enhances robustness against catastrophic forgetting by interpolating between old and new model weights after each novel task, effectively merging two models to facilitate exploration of local minima emerging after arrival of new concepts. Moreover, we demonstrate that our approach can complement existing rehearsal-based replay approaches, improving their accuracy and further mitigating the forgetting phenomenon. Additionally, our method provides an intuitive mechanism for controlling the stability-plasticity trade-off. Experimental results showcase the significant performance enhancement to state-of-the-art experience replay algorithms the proposed weight consolidation approach offers. Our algorithm can be downloaded from \url{https://github.com/jedrzejkozal/weight-interpolation-cl}.
\end{abstract}

\section{Introduction}
\label{sec:intro}

The properties of loss landscape and their effects on training and generalization were objects of study for a long time \cite{DBLP:journals/corr/abs-1712-09913,foret2021sharpnessaware,DBLP:journals/corr/SagunBL16,Sankar2020ADL}. Training of neural network is an optimization process in a highly dimensional non-convex parameter space with many local minima and saddle points \cite{DBLP:journals/corr/PascanuDGB14}. 
Overabundance of local minima may arise due to the overparameterization of neural networks \cite{karhadkar2024mildly}.
It was hypothesized that local minima are connected by non-linear paths with a low loss \cite{garipov2018loss}. This property is known as mode connectivity.
One feature that may be considered when studying this phenomenon is the permutation invariance of neural networks \cite{DBLP:journals/corr/abs-2110-06296}. Neurons or kernels of network layers can be permuted and, if neighboring layers' outputs and inputs are adjusted, one can obtain a solution that has the same properties as the original model but lies in a completely different part of the loss landscape. Considering this fact, one may conclude that the abundance of local minima in the loss landscape of neural networks results from permutation invariance.
In a follow-up work, Ainsworth et al. \cite{ainsworth2023git} showed how to find permutations of weights that allow for a linear interpolation of weights with low or even near zero barriers. They also showed that there exist solutions in the loss landscape that cannot be reached by applying permutation to units of a neural network.

Previous experiments on loss barriers were made mostly with the assumption that networks trained from two independent initializations are in two different local minima and have similar loss values \cite{ainsworth2023git, jordan2022repair}. In the case of continual learning, this assumption cannot be met, as models are subject to forgetting \cite{catastrophic_forgetting} of previously seen data. 
In \cite{marouf2024weighted} weight averaging was proposed to mitigate catastrophic forgetting for pretrained models, however, parameter symmetries were not considered. Similarly, authors of \cite{kuhnel2023bert} utilize weight interpolation to mitigate forgetting in BERT models, but they do not apply weight permutation. Pena et al. \cite{peña2022rebasin} propose a new weight interpolation method based on Sinkhorn differentiation, but continual learning is not their primary focus, and the scope of continuous learning experiments is very limited. Authors of \cite{stoica2024zipit} introduced a new interpolation method that could be used for models trained with disjoint data distributions, however, they do not carry out continual learning evaluation. 

\smallskip
\noindent \textbf{Research goal.} We propose a novel approach to continual learning that combines weight interpolation for better consolidation of the network capabilities before and after new tasks become available, with experience replay for enhanced robustness to catastrophic forgetting.

\smallskip
\noindent \textbf{Motivation.} The impact of loss landscape properties on continual learning is a very important, yet largely unexplored area \cite{Huang:2021, DBLP:journals/corr/abs-2010-04495}. We know that it plays a crucial role in the process of balancing exploration (learning new tasks) and exploitation (retaining previously learned knowledge) \cite{Gu:2022}. Sudden changes in the loss landscape can cause the model to forget previously learned information, while inhibiting the loss adaptation will hinder the accumulation of the new concepts \cite{Olpadkar:2021}. The presence of local minima associated with new tasks can interfere with the optimization process for previous tasks, affecting the model's robustness to catastrophic forgetting \cite{Park:2019}. Therefore, properly understanding and utilizing loss landscape under the continuous nature of data is of vital importance. 

\smallskip
\noindent \textbf{Summary.} In this work, we study the potential applications of recent findings from the field of weight interpolation in continual learning \cite{Chen:2018}. Based on recent weight interpolation techniques, we propose a remarkably simple continual learning algorithm that performs weight interpolation after each task to mitigate forgetting. In this work, we abuse the conjecture about low loss volume being convex modulo permutation symmetries \cite{DBLP:journals/corr/abs-2110-06296}, as each task will have separate data distribution and, consequently, different loss landscapes. However, in our theoretical analysis, we show what conditions should be met to increase the chances of finding good weight permutation and successful interpolation.

We base our approach on widely used experience replay methods. Before training with new data from a new task we store network weights. The training with new data is carried out without any changes from standard replay algorithms. After training we utilize weights trained on the current task and stored old weights to perform permutation and then interpolation. The permutation step aligns units of both networks, while interpolations allow for better knowledge consolidation, compared to replay-based algorithms alone.
We show that our method can reduce forgetting in several rehearsal-based methods. 

\smallskip
\noindent \textbf{Main contributions.} This work offers the following contributions to the continual learning domain:

\begin{itemize}
    \item we show the necessary conditions required for the successful application of weight interpolation to continual learning problems, and verify these claims experimentally;
    \item we propose novel and simple continual learning algorithm that is compatible with popular rehearsal-based methods;
    \item we perform an extensive experimental evaluation of the proposed method, showing its potential for significantly boosting the performance of any experience replay algorithm;
    \item we show that the proposed method has a built-in, intuitive mechanism for controlling stability-plasticity trade-off.
\end{itemize}

\section{Related Works}
\label{sec:related_works}

\subsection{Continual Learning}

Continual learning \cite{Chen:2018} is a domain where, instead of a single i.i.d. dataset, we are dealing with a sequence of tasks with different data distributions. Training without access to data from previous tasks may lead to catastrophic forgetting \cite{catastrophic_forgetting} - a phenomenon where neural network's performance on previous tasks degrades rapidly. The performance here could be defined as losing the ability to solve previously learned tasks when a neural network learns to solve a new one. Catastrophic forgetting could lead to dramatic performance deterioration on the previous tasks.  
In the domain of continual learning, algorithms are typically categorized into three primary groups:

\emph{Regularization-based methods} aim to control forgetting by modifying the learning process. Elastic Weight Consolidation (EWC) \cite{DBLP:journals/corr/KirkpatrickPRVD16} introduces an additional regularization term that constrains the learning of important parameters. Learning without Forgetting (LwF) \cite{DBLP:journals/corr/LiH16e} leverages pseudo-labels derived from classification heads of previous tasks to enhance knowledge retention.
Synaptic intelligence \cite{DBLP:journals/corr/ZenkePG17} is a structural regularizer that enforces penalty on each synapse based on its importance for previous tasks.

\emph{Rehearsal-based methods} rely on memory buffers to store samples from previous tasks \cite{DBLP:journals/corr/abs-1902-10486}. Gradient Episodic Memory (GEM) \cite{DBLP:journals/corr/Lopez-PazR17} utilizes examples from memory to project gradients in directions that minimize loss for previous tasks. Averaged GEM (aGEM) \cite{DBLP:journals/corr/abs-1812-00420} is a refined version that offers computational and memory efficiency. Recent investigations \cite{chrysakis2023online} have explored asymmetric update rules and additional classifier updates to address biases introduced by small rehearsal buffers.
Moreover, Buzzega et al. \cite{buzzega2020dark} stores model logits alongside images and labels in a memory buffer. These logits are subsequently utilized to regulate the model by introducing an additional loss term for knowledge distillation. This method was refined in \cite{DBLP:journals/corr/abs-2201-00766} by recalculating logits over time, segregating loss for new data, and pretraining logits responsible for new tasks.
There are also other research directions, such as iCARL \cite{DBLP:journals/corr/RebuffiKL16}, where instead of cross-entropy loss, a minimal distance classifier is trained on top of a convolutional neural network. 
Another interesting and simple algorithm is GDumb \cite{10.1007/978-3-030-58536-5_31}, which utilizes only greedily stored samples to train the model with the small balanced dataset. 

\emph{Expansion-based methods} involve augmenting the network structure to accommodate shifts in data distribution. Progressive Neural Networks (PNN) \cite{DBLP:journals/corr/RusuRDSKKPH16} add new backbones connected to previous layers to leverage knowledge learned from earlier tasks. Similarly, \cite{DBLP:journals/corr/abs-1708-01547} propose expanding network parameters alongside selective retraining to adapt to new tasks.
Authors of \cite{DBLP:journals/corr/abs-2103-16788} expand the model by introducing more convolutional features for new tasks, and they propose a new loss function to train a more diverse set of representations for new data.

\subsection{Weight interpolation}

Garipov et al. \cite{garipov2018loss} showed that local minima obtained by training with different random weight initialization in the loss landscape are connected by non-linear paths with low loss values. This property was introduced as \emph{Mode Connectivity}. 
It is also known \cite{DBLP:journals/corr/abs-1907-02911} that there exists a lot of possible weight permutations that give raise to equivalent networks located in completely different fragments of loss landscape.
In \cite{DBLP:journals/corr/abs-2009-02439}, a new algorithm for finding network permutations and the connection curve between two points in the loss landscape was introduced. Authors of \cite{DBLP:journals/corr/abs-1902-04742} showed that weights of MLP trained from the same initialization can be linearly connected.
Entezari et al. \cite{DBLP:journals/corr/abs-2110-06296} suggested that when we consider neural network permutation invariance, solutions found by SGD should be connected by linear path no loss barrier. Indeed, Ainsworth et al. \cite{ainsworth2023git} proposed several algorithms for finding permutations of neural networks that allow for linear interpolation between weights with near-zero barrier. 
REPAIR \cite{jordan2022repair} improved the performance for residual networks on bigger datasets by introducing the recomputation of batch normalization statistics after interpolation.


\section{Continual learning with weight interpolation}
\label{sec:clewi}

\begin{figure*}
    \centering
    \resizebox{10cm}{!}{
    \includegraphics[width=1.0\linewidth]{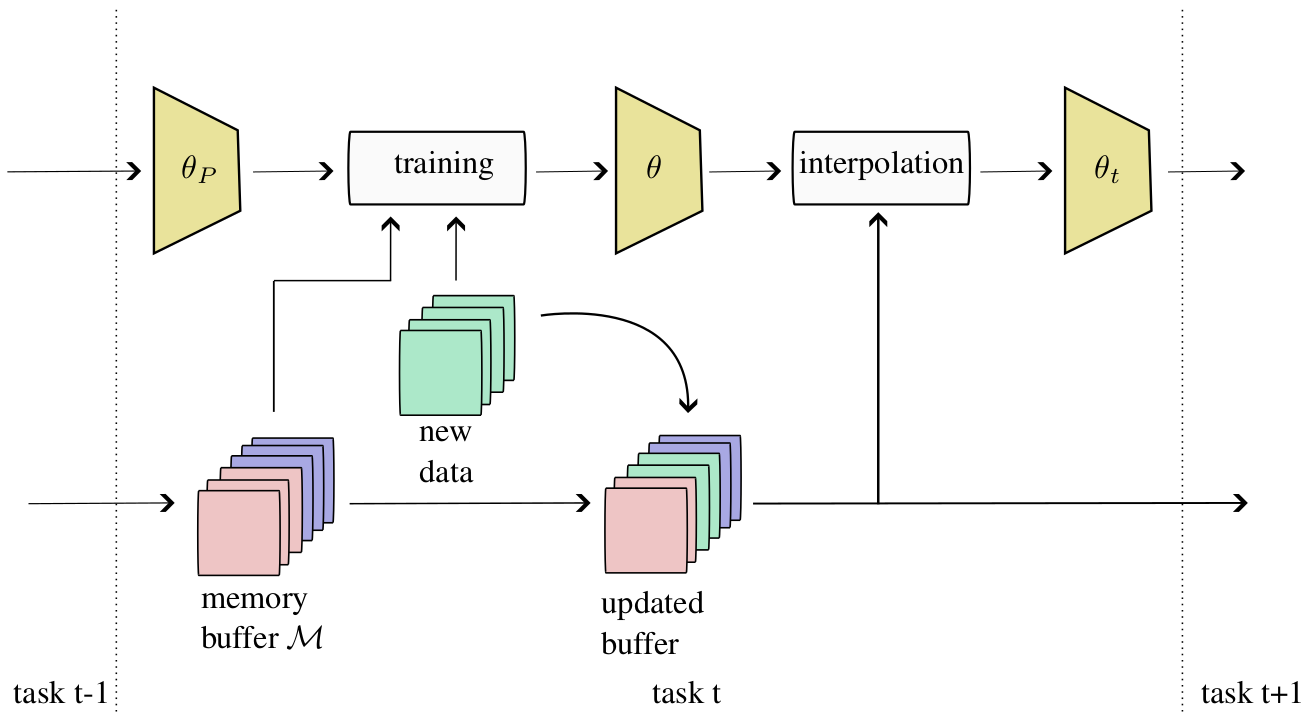}
    }
    \caption{Continual learning with weight interpolation.}
    \label{fig:method-schema}
\end{figure*}

This work introduces a simple method that could be used as a plugin to enhance the effectiveness of any rehearsal algorithm. The core idea is to interpolate weights of a neural network before and after training with new data. This should allow for better knowledge consolidation and inhibit forgetting. An overview of the proposed method is provided in \cref{fig:method-schema}.

\subsection{Notation}

In continual learning, we are dealing with stream $S$, arriving in the form of tasks. Each task $t$ may be represented by a dataset $D_t = \{ (x_i, y_i) \}_{i=0}^{n_t}$, where $x_i$ is image, $y_i$ is label, and $n_t=|D_t|$. 
The goal is to train a neural network $f$ with parameters $\theta$ on each task, having access only to the most recent data, i.e., $\min_{\theta} \mathcal{L}(f(\theta), D_t)$.
Rehearsal-based algorithms utilize an additional small buffer for data $\mathcal{M} = \{ (x_j, y_j) \}_{j=0}^{m}$ of size $m \ll n_t$ to store data from the previous task and use them to mitigate forgetting. 

\subsection{Motivation}

The main objective of continual learning is the optimization of the joint test loss across all tasks in the stream. We can only access the training data from the current task, but our main goal is to train the network with a low loss across all tasks.
Joint loss for all tasks seen so far by the model can be defined as: 

\begin{equation}
    \mathcal{L}_{D}(\theta) = \sum_{t=1}^{T} \mathcal{L}(\theta, D_t)
\end{equation}

\noindent where $D = D_1\cup \dots \cup D_T$. We can divide this sum into two parts, namely, loss induced by the classes from the last task and all other classes seen before:

\begin{equation}
    \mathcal{L}_{D}(\theta) = \sum_{t=1}^{T-1} \mathcal{L}(\theta, D_t) + \mathcal{L}(\theta, D_T)
    \label{eq:L_C}
\end{equation}

The first term corresponds to performance on all previous tasks and is mainly affected by forgetting in a continual learning setup. The second term can be directly optimized for, as we have access to data for the task $T$. 
Let's define an increase in loss induced by forgetting tasks before $T$ as:

\begin{equation}
    \Delta \mathcal{L}_{Fi} = \sum_{t=1}^{i-1} (\mathcal{L}(\theta_i, D_t) - \mathcal{L}(\theta_t, D_t))
    \label{eq:delta_L}
\end{equation}

The first term inside the sum is the current loss for task $i$, and the second term is the loss directly after training with data from the same task. This definition is analogous to forgetting measure \cite{DBLP:journals/corr/abs-1801-10112} - a commonly used metric designed for evaluation of accuracy decrease during continual training. By plugging \cref{eq:delta_L} into \cref{eq:L_C} we can rewrite joint loss function after task $T$ as:

\begin{equation}
    \mathcal{L}_D(\theta_T) = \sum_{t=1}^{T} \mathcal{L}(\theta_t, D_t) + \Delta \mathcal{L}_{Fi=T}
    \label{eq:limit1}
\end{equation}

Therefore, loss obtained by the network depends on two factors: (i) how well the network can fit current data, which corresponds to plasticity; and (ii) how well the network handles the previously seen data, which corresponds to forgetting.
The second network used for interpolation is the one trained on the previous task $T-1$. We can make the same argument about loss being dependent on plasticity and forgetting/ 

When we search for good candidates for interpolation between $\theta_{T-1}$ and $\theta_T$ we require both $\mathcal{L}_D(\theta_T)$ and $\mathcal{L}_D(\theta_{T-1})$ to be low.
This is because we must have solutions either in the local minima or close to some local basin. 
As shown by \cref{eq:limit1}, this can achieved only when both plasticity is high and forgetting is low. If that is not the case, then the loss term induced by any of these terms could increase the overall loss value, moving away the solution in the loss landscape from the locally connected modes.
Interpolation with weight permutation alone can, in principle, align the activations of the networks trained on the different tasks, so there could be a gain in accuracy directly after interpolation. Still, if the activations learned by the network on the new task are completely different from the previous ones due to forgetting, then alignment between activations can be inaccurate. On the other hand, if there is no plasticity, then alignment could be easier, but there would be no significant difference between the two sets of activations.

For this reason, we conclude that weight interpolation should not be used as a sole source of forgetting prevention in continual learning.
Weight interpolation could be used in tandem with other continual learning algorithms that do not limit network plasticity too much. To further justify this claim, we verify experimentally in the appendix \ref{sec:clewi_without_rehearsal} that using interpolation without rehearsal does not yield good results.



\subsection{Weight interpolation with memory buffer}

For each task $t > 0$, we perform weight interpolation of previously trained weights $\theta_P$ with the newest parameters trained with current data distribution $\theta$. First, we find the weight permutation $\pi$ that aligns the activations of $\theta_P$ and $\theta$ as in \cite{jordan2022repair} (for more details about interpolation and REPAIR algorithms, please refer to \cref{sec:repair}). We utilize memory buffer $\mathcal{M}$ to obtain activations of $\theta$ and $\theta_P$ and update batch normalization statistics. Please note that if we use reservoir sampling during training to update the buffer with new data, the buffer will contain the data from all previous and current tasks. For this reason, during the evaluation of activations for permutation, all previously seen data will be considered, including data from the latest task.
We apply the permutation to network parameters and carry out linear interpolation of weights:

\begin{equation}
    \theta = (1-\alpha) \theta + \alpha \pi(\theta_P)
    \label{eq:interpolation}
\end{equation}

\noindent where $\alpha$ is a hyperparameter of our algorithm. 
We provide the pseudocode of our method in~\cref{alg:clewi}. The function $\textit{calc\_permutation}$ is responsible for obtaining permutation of $\theta_P$ that aligns activations of $\theta_P$ with $\theta$. The function $\textit{update\_batchnorm}$ updates the batch normalization layers statistics after interpolation. 
The proposed method is compatible with most rehearsal-based algorithms and may be used as a plugin for existing or future methods for improving their performance.

\begin{algorithm}[t]
\caption{Continual Learning with Weight Interpolation (CLeWI)}
\label{alg:clewi}
\begin{algorithmic}[1]
\Require $S = \{ D_1, D_2,... \} $ - stream with tasks, $f(\theta)$ - network, $\mathcal{M}$ - memory buffer, $\alpha$ - interpolation coefficient
\State $t \gets 0$
\While {$D_t$ arrives} 
    \For{$x,y \sim D_t$}
        \State $\mathcal{L} \gets \sum_{x, y} \mathcal{L}(f(x, \theta), y)$
        \State $x_m, y_m \gets \mathcal{M}$
        \State $\mathcal{L_M} \gets \sum_{x_m, y_m} \mathcal{L}(f(x_m, \theta), y_m)$
        \State $\theta \gets \theta - \lambda \nabla_{\theta} (\mathcal{L} + \mathcal{L_M})$
        \State \textit{resevoir\_sampling} ($\mathcal{M}, x, y$)
    \EndFor
    \If{$t > 0$}
        \State $\pi \gets \textit{calc\_permutation}(\theta, \theta_P, \mathcal{M})$
        \State $\theta \gets (1-\alpha) \theta + \alpha\pi(\theta_P)$
        \State $\theta \gets \textit{update\_batchnorm}(\theta, \mathcal{M})$
    \EndIf
    \State $\theta_P \gets \theta$
    \State $t \gets t + 1$
\EndWhile
\end{algorithmic}
\end{algorithm}

\section{Experiment setup}
\label{sec:experiments_setup}

Our experiments compare the performance of commonly used rehearsal algorithms with and without weight interpolation applied after each task. This evaluation mode, similar to the ablation study, should allow for an easy verification of our theoretical claims made in the previous section. We also provide in-depth analysis of the weight interpolation impact on the overall performance, and stability-plasticity dilemma. We also evaluate the impact of the training with increased model width on the results.

\noindent \textbf{Baselines.} In this work, we have used the following baselines:

\begin{itemize}
    \item joint - training with cumulative datasets over all tasks, with full access to previous data. It is upperbound on the continual learning performance.
    \item finetuning - training with standard SGD optimization, with no consideration for forgetting. It is lowerbound of performance
    \item online Elastic Weight Consolidation (oEWC) \cite{schwarz2018progress} - an extension of existing EWC method \cite{DBLP:journals/corr/KirkpatrickPRVD16}, that use both regularisation and knowledge distillation to prevent forgetting.
    \item Synaptic Inteligence \cite{pmlr-v70-zenke17a} - regularisation method that determines the importance of network parameters.
    \item Incremental Classifier and Representation Learning (iCARL) \cite{DBLP:journals/corr/RebuffiKL16} - method that replaces cross entropy with prototype-based learning
    \item GDumb \cite{prabhu2020gdumb} - Greedily stores samples in memory and trains model only with balanced dataset
    \item Experience Replay (ER) \cite{DBLP:journals/corr/abs-1902-10486} - simplest rehearsal method that stores samples in the buffer using reservoir sampling and samples data from the buffer to train with it alongside data new from a new task
    \item averaged Gradient Episodic Memory (aGEM) \cite{DBLP:journals/corr/abs-1812-00420} rehearsal method, that projects gradient onto direction, that prevents forgetting
    \item  Experience Replay with Asymmetric Cross-Entrop (ER-ACE) \cite{DBLP:journals/corr/abs-2104-05025} - eliminates representation overlap of new classes and old ones from the buffer by changing the loss function
    \item Maximally Interfered Retrieval (MIR) \cite{DBLP:journals/corr/abs-1908-04742} - method with the buffer that uses virtual gradient update to select useful samples for rehearsing
    \item Bias Correction (BIC) \cite{Wu2019LargeSI} - a method that introduces several parameters for correction of bias in the last fully connected layer of the network
    \item Dark Experience Replay (DER++) \cite{buzzega2020dark} - method that combines rehearsal with knowledge distillation \cite{hinton2015distilling}
\end{itemize}

\smallskip
\noindent \textbf{Datasets.} In this work, we consider only the class-incremental scenario \cite{DBLP:journals/corr/abs-1904-07734} and utilize standard continual learning benchmarks obtained by splitting classes into several tasks. We use Cifar10 \cite{Krizhevsky09learningmultiple}, Cifar100 \cite{Krizhevsky09learningmultiple}, and Tiny ImageNet \cite{Wu2017TinyIC} datasets, with 5, 10, and 20 tasks, respectively. We shuffle class order in tasks based on random seeds. 

\smallskip
\noindent \textbf{Metrics.} We use three evaluation metrics. The test set accuracy averaged over all tasks after finished training, defined as $Acc = \frac{1}{K} \sum_t^K \frac{1}{n_t} \sum_{i=1}^{n_t} \mathbb{1}[f(x_i, \theta_K) = y_i]$, where $K$ is the number of tasks, and $\mathbb{1}$ is an indicator function. 
The test set accuracy for classes from the last task $Acc_K = 
\frac{1}{n_K} \sum_{(x_i, y_i) \in D_K} \mathbb{1}[f(x_i, \theta_K) = y_i]$, and
forgetting measure (FM) \cite{DBLP:journals/corr/abs-1801-10112} defined as average difference between maximum accuracy, and final accuracy for given task.

\smallskip
\noindent \textbf{Evaluation details.} For all datasets, we use ResNet18 architecture \cite{DBLP:journals/corr/HeZRS15} with a changed number of filters in the first layer following \cite{DBLP:journals/corr/Lopez-PazR17}.
For all rehearsal-based methods, we use a buffer of size 500.
Whenever possible, we use the best hyperparameters reported by the authors of corresponding papers. In other cases, we performed a search of hyperparameters for the seq-cifar100 benchmark and used those values for other datasets as well. This shortcut has been made due to limitations in computational power availability.
All experiments were implemented using Mammoth library \cite{buzzega2020dark}. 
We made our code available online\footnote{\url{https://github.com/jedrzejkozal/weight-interpolation-cl}}.

\section{Results}
\label{sec:results}

\subsection{Evaluation with standard benchmarks}

\begin{table*}
    \centering
    \scriptsize
    \caption{Average accuracy and forgetting measure averaged over 5 runs for cifar10, cifar100, and tinyimagenet datasets.}

\begin{tabular}{c|cc|cc|cc}
    \hline
    \multirow{2}{*}{method} & \multicolumn{2}{c}{Cifar10(T=5)} & \multicolumn{2}{c}{Cifar100(T=10)}  & \multicolumn{2}{c}{Tiny-ImageNet(T=20)} \\
    \cline{2-7} 
     & Acc($\uparrow$) & FM$(\downarrow)$ & Acc($\uparrow$) & FM$(\downarrow)$ & Acc($\uparrow$) & FM$(\downarrow)$ \\
    \hline
 Joint        & 91.79±0.36                            & 0.0±0.0                              & 70.54±0.75                            & 0.0±0.0                               & 58.34±0.24                            & 0.0±0.0                               \\
 Finetuning   & 19.37±0.32                            & 77.77±0.85                           & 9.07±0.1                              & 80.57±0.41                            & 3.92±0.27                             & 74.75±1.34                            \\
 oEWC         & 17.21±2.89                            & 69.94±3.98                           & 8.86±0.51                             & 76.05±0.43                            & 3.71±0.23                             & 70.14±1.6                             \\
 SI           & 19.28±0.4                             & 78.11±0.38                           & 6.36±0.53                             & 36.99±1.32                            & 3.64±0.4                              & 67.97±2.1                             \\
 iCARL        & \underline{58.98±1.21}                            & 25.27±4.72                           & \underline{46.91±0.66}                            & 25.56±0.57                            & \underline{19.69±0.37}                            & 20.24±0.54                            \\
 GDumb        & 39.7±1.57                             & \underline{0.66±0.65}                            & 9.99±0.68                             & \underline{0.0±0.0}                               & 3.2±0.31                              & \underline{0.22±0.15}                             \\
\hline
 ER           & 53.22±2.98                            & 44.02±3.59                           & 22.45±1.26                            & 65.59±1.07                            & 6.44±0.38                             & 75.88±0.23                            \\
 CLeWI+ER     & 62.8±2.31(\textcolor{Green}{+9.58})   & 31.8±2.61(\textcolor{Green}{-12.22}) & \underline{40.31±1.08}(\textcolor{Green}{+17.86}) & \underline{12.81±0.79}(\textcolor{Green}{-52.78}) & 11.68±0.45(\textcolor{Green}{+5.24})  & 66.82±0.49(\textcolor{Green}{-9.06})  \\
 aGEM         & 21.88±1.15                            & 75.63±0.96                           & 9.17±0.18                             & 80.33±0.34                            & 3.62±0.54                             & 73.61±3.29                            \\
 CLeWI+aGEM   & 34.74±4.05(\textcolor{Green}{+12.86}) & 4.16±1.92(\textcolor{Green}{-71.47}) & 22.75±1.41(\textcolor{Green}{+13.58}) & 39.07±2.26(\textcolor{Green}{-41.26}) & 6.8±0.4(\textcolor{Green}{+3.18})     & 60.22±1.17(\textcolor{Green}{-13.39}) \\
 ER-ACE       & 70.63±1.15                            & 10.11±0.95                           & 37.75±1.23                            & 35.15±1.33                            & 15.98±1.64                            & 42.47±2.43                            \\
 CLeWI+ER-ACE & 64.22±2.5(\textcolor{Red}{-6.41})     & \underline{4.84±0.67}(\textcolor{Green}{-5.27})  & 36.97±0.55(\textcolor{Red}{-0.78})    & 17.72±0.76(\textcolor{Green}{-17.43}) & 19.15±0.72(\textcolor{Green}{+3.17})  & \underline{19.7±0.88}(\textcolor{Green}{-22.77})  \\
 MIR          & 48.17±3.23                            & 49.02±3.72                           & 21.96±1.13                            & 66.07±0.95                            & 6.25±0.41                             & 76.06±0.3                             \\
 CLeWI+MIR    & \underline{73.06±0.74}(\textcolor{Green}{+24.89}) & 6.71±0.84(\textcolor{Green}{-42.31}) & 40.06±0.84(\textcolor{Green}{+18.10}) & 13.54±0.54(\textcolor{Green}{-52.53}) & \underline{19.75±0.56}(\textcolor{Green}{+13.50}) & 25.47±0.43(\textcolor{Green}{-50.59}) \\
 BIC          & 69.63±2.28                            & 22.04±3.04                           & 37.55±1.64                            & 44.42±1.87                            & 7.09±0.78                             & 71.47±0.87                            \\
 CLeWI+BIC    & 51.15±9.53(\textcolor{Red}{-18.48})   & 26.48±4.96(\textcolor{Red}{+4.44})   & 39.46±1.34(\textcolor{Green}{+1.91})  & 33.46±1.46(\textcolor{Green}{-10.96}) & 7.35±1.4(\textcolor{Green}{+0.26})    & 65.34±0.79(\textcolor{Green}{-6.13})  \\
 DER++        & 70.13±1.16                            & 21.11±1.46                           & 36.64±1.59                            & 48.06±2.62                            & 13.52±1.53                            & 55.68±4.36                            \\
 CLeWI+DER++  & 71.82±2.11(\textcolor{Green}{+1.69})  & 11.21±2.16(\textcolor{Green}{-9.90}) & 38.16±1.86(\textcolor{Green}{+1.52})  & 14.32±1.95(\textcolor{Green}{-33.74}) & 16.61±0.87(\textcolor{Green}{+3.09})  & 25.17±6.34(\textcolor{Green}{-30.51}) \\
\hline
\end{tabular}
    \label{tab:eval_bechmarks}
\end{table*}

We perform an experimental evaluation of the proposed method, following the steup described in the previous section. The results are presented in ~\cref{tab:eval_bechmarks}.

In most cases, we may see that the proposed method improves the average accuracy on all tasks and leads to better task retention, as depicted by reducing the forgetting measure. 
The biggest gains in accuracy can be observed for simpler forms of replay, such as ER and MIR. CLeWI obtains the best accuracy when combined with these methods.
Other methods, such as ER-ACE, BIC, or DER++, can also benefit from applying interpolation. However, the final average accuracy after training is lower compared to simpler methods. At the same time, the forgetting rate of these methods is lower than that of others.
These methods limit the plasticity of the networks. In the interpolation process, we are losing some of the performance for the newest task at the cost of forgetting mitigation. For this reason, after applying interpolation, the methods that obtain lower forgetting on their own can sometimes obtain lower accuracy and forgetting measure. These results are in line with our theoretical analysis. Low plasticity can contribute to high overall loss and, in consequence, make interpolation harder.

We noted a decrease in performance for ER-ACE, where average accuracy is lower, but forgetting measure still improves, and BIC on the Cifar100 dataset, where both accuracy and forgetting measure are worse. This is in line with our previous analysis, as these methods introduce strong inductive bias and obtain higher accuracy compared to ER. CLeWI, when combined with these methods, inherits this bias, and therefore, average accuracy can decrease.




\subsection{Impact of weight interpolation}

\begin{figure}
    \centering
    \includegraphics[width=1.0\linewidth]{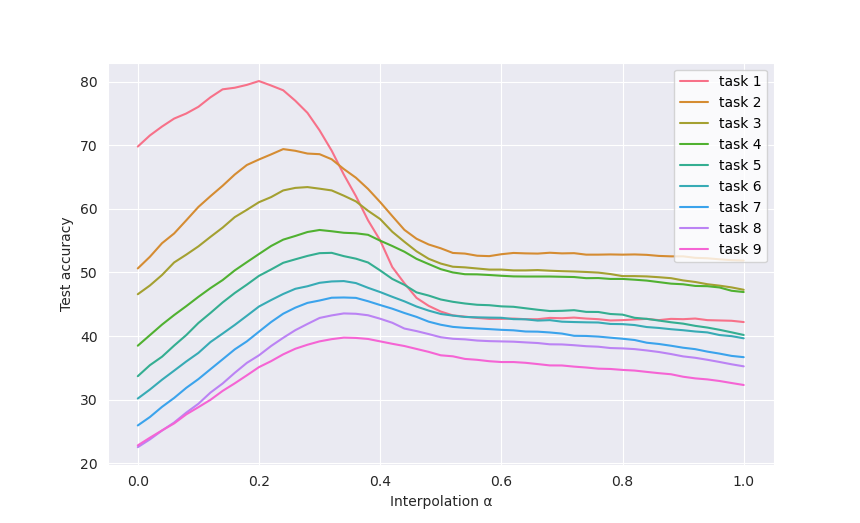}
    \caption{The effect of the $\alpha$ parameter (\cref{eq:interpolation}) on the test set accuracy for all tasks. Interpolation with smaller values of $\alpha$ allows for obtaining weights that are closer in loss landscape to the current task, while increasing $\alpha$ means more weights are carried over from previous tasks.
    }
    \label{fig:interpolation}
\end{figure}

To illustrate the influence of interpolation hyperparameter $\alpha$ on obtained results we plot accuracy for different interpolation $\alpha$ and all tasks. We use all the classes the model has seen for each task. This means that the older model will always obtain worse performance, as it has not seen the classes from the latest task. Results are presented in \cref{fig:interpolation}. 
At the beginning of training, better overall accuracy is obtained after training with a new task compared to the model weights before training. This is probably due to underfitting on the first tasks caused by a small number of learning examples in each task. Over the course of training, the difference in accuracy between these two models falls quickly. After a few tasks, the old model performs better, while the new one suffers from forgetting. The interpolation plot for continual learning is asymmetrical. 
Interpolating models closer to the model trained on a new task gives better accuracy. This is probably due to the longer training of the model with data from the new task.

\subsection{Stability-plastisity dilemma}
We show that interpolation hyperparameter $\alpha$ allows for direct control of the plasticity-stability dilemma by running additional experiments with multiple values of this hyperparameter. The results are presented in \cref{tab:interpolation_alpha}. With higher $\alpha$ (interpolation closer to the old model), the model is prone to remembering the older tasks. This can be directly observed by looking at forgetting measures. Higher $\alpha$ usage promotes stability and limits performance for the current task.
With smaller $\alpha$ (interpolation closer to a newer model), the network archives better accuracy on the last task at the price of higher forgetting.
This simple mechanism could be useful for controlling the learning properties of neural networks. Interpolation $\alpha$ may also be changed during training with multiple tasks to adapt to the changing dynamics of the learning environment. 

\begin{table}
    \scriptsize
    \centering
    \caption{Average accuracy, accuracy for the last task, and forgetting measure averaged over 3 runs for different values of interpolation $\alpha$.}
    \begin{tabular}{c|c|c|c}
    \hline
    \makecell{interpolation\\coefficient} & Acc($\uparrow$) & Acc$_K$($\uparrow$) & FM($\downarrow$)  \\
    \hline
 $\alpha$=0.1 & 27.6±0.46  & 87.77±2.09 & 59.11±0.66 \\
 $\alpha$=0.2 & 34.75±0.51 & 83.87±2.57 & 47.14±0.23 \\
 $\alpha$=0.3 & 39.95±0.67 & 72.23±3.32 & 30.67±0.55 \\
 $\alpha$=0.4 & 42.01±0.82 & 44.6±4.78  & 18.9±0.24  \\
 $\alpha$=0.5 & 40.26±1.25 & 16.27±4.47 & 12.61±0.96 \\
 \hline
\end{tabular}
    \label{tab:interpolation_alpha}
\end{table}

Comparing these results to \cref{fig:interpolation}, one can notice that the best test accuracy was obtained for a value of $\alpha$ that is not aligned with the local maximum of the interpolation plot. This suggests that selecting $\alpha$ only to optimize the performance on tasks seen so far is a misleading approach that can lead to lower accuracy at the end of training.

\subsection{Wider networks}

\begin{figure}
    \centering
    \includegraphics[width=0.48\linewidth]{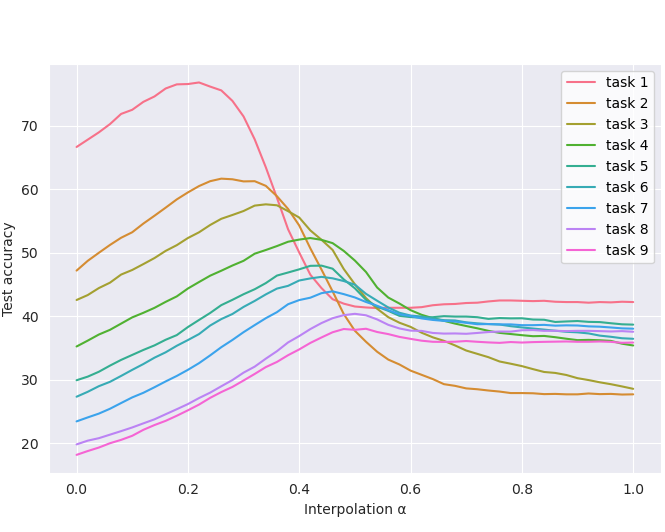}
    \includegraphics[width=0.48\linewidth]{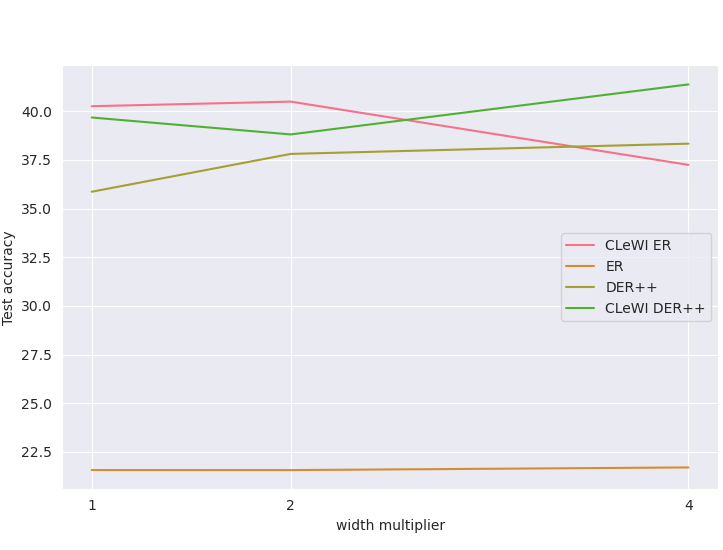}
    \caption{Impact of increasing the network width on the accuracy barrier and continual learning performance. (Left) the interpolation plot for the WideResNet with width multiplier $=4$. (Right) test accuracy for split-Cifar100 benchmark as a function of ResNet width.} 
    \label{fig:width}
\end{figure}

It has been reported that interpolation works better when network architecture has more filters \cite{ainsworth2023git, jordan2022repair}. Also, recent studies suggest that wider architectures could lead to improved performance in continual learning \cite{DBLP:journals/corr/abs-2202-00275}. For this reason, we have carried out additional experiments with WideResNets \cite{DBLP:journals/corr/ZagoruykoK16} on the split-Cifar100 benchmark. We kept the same hyperparameter setting, only the width was changed. The results are presented in \cref{fig:width}

On the left-hand side, we present the interpolation plot for the WideResNet with an increased number of convolutional filters by 4. We may see that compared to results from \cref{fig:interpolation}, the localization of the local maximum accuracy shifts more dynamically during training. 
It is also worth noting that in \cref{fig:interpolation}, the accuracy for $\alpha$ close to 1 decreases slightly. This is not the case for a wider network, where the accuracy on the plot is mostly flat for $\alpha > 0.7$. This suggests that wider networks are indeed better at preserving previously gained knowledge. However, this does not translate well into overall continual learning performance when using CLeWI due to dynamic changes in the shape of interpolation plot curves for different tasks. We hypothesize that these changes arise due to the small amount of training data for WideResNet in a single task of Cifar100 benchmark. However, we are aware that experiments with other benchmarks, such as split-ImageNet, could provide different results and further investigations are needed.

The right-hand side shows the test accuracy in the function of ResNet width. We may see that increasing the width could significantly improve the performance of DER++, but even with this improvement, CLEWI-DER++ with standard width obtains better performance. 
At the same time, increasing the width of the backbone for the CLeWI ER decreases accuracy. 
The small amount of training data in each task may be the cause behind this phenomenon. Overfitting may occur when we increase the network's capacity but keep the same amount of training data. The dynamic changes in local maxima of the interpolation plots for increased width are in line with this explanation. For the first task, local maxima are obtained for smaller $\alpha$ - corresponding to interpolation closer to the newer model. All these information suggest that experience replay alone is insufficient when training networks with larger capacity.
When stronger forgetting prevention mechanisms are introduced, such as DER++, the performance of CLeWI further improves with increased width. This shows that our algorithm is a versatile approach to boosting the performance of CL methods due to the ease of combining CLeWI with other forms of rehearsal. The proposed method can be easily adjusted to other settings by combining it with the form of rehearsal that works well in a given scenario. This shows the flexibility of CLeWI and its strength as a low-risk, low-cost plugin for existing methods.

\section{Conclusion}
\label{sec:conclusion}

\noindent \textbf{Summary.} We proposed a simple algorithm, compatible with most of the rehearsal-based continual learning methods that can significantly boosts their performance and improve robustness to catastrophic forgetting. CLeWI introduces only a single additional hyperparameter that allows for direct control of the stability-plasticity dilemma. 
The experiments suggested that $\alpha$ selection should be carried out with great care, as local maxima for the current task not necessarily align well with higher accuracy for all tasks in the training stream. In the interpolation plots, we may see that local maxima's location can shift over time. In earlier tasks, the maxima occur for lower values of $\alpha$, probably due to too small amount of training data in each task. Experiments with bigger datasets could provide more insight here, as we hypothesise, that with enough data in the first task, the location of local minima in the interpolation plot will be more stable.

\smallskip
\noindent \textbf{Limitations.} Storing a second copy of model weights in memory can be prohibitive for large models. For example, when training 1.4B parameter transformer storing previous model state could be too costly. Additional memory requirements may also be prohibitive in the memory-scarce area of edge computing. 
We carried out additional experiments (see \cref{sec:memory_restricted} in appendix) that take into consideration memory usage. 
We found settings where using weight interpolation over increasing buffer size alone can be beneficial.

\smallskip
\noindent \textbf{Future works.} Future work will focus on exploring of the weight interpolation should be performed after every new task, or would a selective mechanism deciding when to perform interpolation lead to more robust results. Furthermore, we will explore the potential of using CLeWI as a part of concept drift adaptation mechanisms \cite{Korycki:2021} and study the possibilities of extending it for other computer vision tasks, such as object detection or continual segmentation.

\section*{Acknowledgment}

This work is supported by the CEUS-UNISONO programme,
which has received funding from the National Science Centre,
Poland under grant agreement No. 2020/02/Y/ST6/00037.

{
    \small
    \bibliographystyle{ieeenat_fullname}
    \bibliography{main}

\begin{thebibliography}{57}
\providecommand{\natexlab}[1]{#1}
\providecommand{\url}[1]{\texttt{#1}}
\expandafter\ifx\csname urlstyle\endcsname\relax
  \providecommand{\doi}[1]{doi: #1}\else
  \providecommand{\doi}{doi: \begingroup \urlstyle{rm}\Url}\fi

\bibitem[Ainsworth et~al.(2023)Ainsworth, Hayase, and
  Srinivasa]{ainsworth2023git}
Samuel~K. Ainsworth, Jonathan Hayase, and Siddhartha Srinivasa.
\newblock Git re-basin: Merging models modulo permutation symmetries, 2023.

\bibitem[Aljundi et~al.(2019)Aljundi, Caccia, Belilovsky, Caccia, Lin, Charlin,
  and Tuytelaars]{DBLP:journals/corr/abs-1908-04742}
Rahaf Aljundi, Lucas Caccia, Eugene Belilovsky, Massimo Caccia, Min Lin,
  Laurent Charlin, and Tinne Tuytelaars.
\newblock Online continual learning with maximally interfered retrieval.
\newblock \emph{CoRR}, abs/1908.04742, 2019.

\bibitem[Boschini et~al.(2022)Boschini, Bonicelli, Buzzega, Porrello, and
  Calderara]{DBLP:journals/corr/abs-2201-00766}
Matteo Boschini, Lorenzo Bonicelli, Pietro Buzzega, Angelo Porrello, and Simone
  Calderara.
\newblock Class-incremental continual learning into the extended der-verse.
\newblock \emph{CoRR}, abs/2201.00766, 2022.

\bibitem[Brea et~al.(2019)Brea, Simsek, Illing, and
  Gerstner]{DBLP:journals/corr/abs-1907-02911}
Johanni Brea, Berfin Simsek, Bernd Illing, and Wulfram Gerstner.
\newblock Weight-space symmetry in deep networks gives rise to permutation
  saddles, connected by equal-loss valleys across the loss landscape.
\newblock \emph{CoRR}, abs/1907.02911, 2019.

\bibitem[Buzzega et~al.(2020)Buzzega, Boschini, Porrello, Abati, and
  Calderara]{buzzega2020dark}
Pietro Buzzega, Matteo Boschini, Angelo Porrello, Davide Abati, and Simone
  Calderara.
\newblock Dark experience for general continual learning: a strong, simple
  baseline.
\newblock In \emph{Advances in Neural Information Processing Systems}, pages
  15920--15930. Curran Associates, Inc., 2020.

\bibitem[Caccia et~al.(2021)Caccia, Aljundi, Tuytelaars, Pineau, and
  Belilovsky]{DBLP:journals/corr/abs-2104-05025}
Lucas Caccia, Rahaf Aljundi, Tinne Tuytelaars, Joelle Pineau, and Eugene
  Belilovsky.
\newblock Reducing representation drift in online continual learning.
\newblock \emph{CoRR}, abs/2104.05025, 2021.

\bibitem[Chaudhry et~al.(2018{\natexlab{a}})Chaudhry, Dokania, Ajanthan, and
  Torr]{DBLP:journals/corr/abs-1801-10112}
Arslan Chaudhry, Puneet~Kumar Dokania, Thalaiyasingam Ajanthan, and Philip
  H.~S. Torr.
\newblock Riemannian walk for incremental learning: Understanding forgetting
  and intransigence.
\newblock \emph{CoRR}, abs/1801.10112, 2018{\natexlab{a}}.

\bibitem[Chaudhry et~al.(2018{\natexlab{b}})Chaudhry, Ranzato, Rohrbach, and
  Elhoseiny]{DBLP:journals/corr/abs-1812-00420}
Arslan Chaudhry, Marc'Aurelio Ranzato, Marcus Rohrbach, and Mohamed Elhoseiny.
\newblock Efficient lifelong learning with {A-GEM}.
\newblock \emph{CoRR}, abs/1812.00420, 2018{\natexlab{b}}.

\bibitem[Chaudhry et~al.(2019)Chaudhry, Rohrbach, Elhoseiny, Ajanthan, Dokania,
  Torr, and Ranzato]{DBLP:journals/corr/abs-1902-10486}
Arslan Chaudhry, Marcus Rohrbach, Mohamed Elhoseiny, Thalaiyasingam Ajanthan,
  Puneet~Kumar Dokania, Philip H.~S. Torr, and Marc'Aurelio Ranzato.
\newblock Continual learning with tiny episodic memories.
\newblock \emph{CoRR}, abs/1902.10486, 2019.

\bibitem[Chen et~al.(2018)Chen, Liu, Brachman, Stone, and Rossi]{Chen:2018}
Zhiyuan Chen, Bing Liu, Ronald Brachman, Peter Stone, and Francesca Rossi.
\newblock \emph{Lifelong Machine Learning}.
\newblock Morgan {\&} Claypool Publishers, 2nd edition, 2018.

\bibitem[Chrysakis and Moens(2023)]{chrysakis2023online}
Aristotelis Chrysakis and Marie-Francine Moens.
\newblock Online bias correction for task-free continual learning.
\newblock In \emph{The Eleventh International Conference on Learning
  Representations}, 2023.

\bibitem[Entezari et~al.(2022)Entezari, Sedghi, Saukh, and
  Neyshabur]{DBLP:journals/corr/abs-2110-06296}
Rahim Entezari, Hanie Sedghi, Olga Saukh, and Behnam Neyshabur.
\newblock The role of permutation invariance in linear mode connectivity of
  neural networks.
\newblock In \emph{International Conference on Learning Representations}, 2022.

\bibitem[Foret et~al.(2021)Foret, Kleiner, Mobahi, and
  Neyshabur]{foret2021sharpnessaware}
Pierre Foret, Ariel Kleiner, Hossein Mobahi, and Behnam Neyshabur.
\newblock Sharpness-aware minimization for efficiently improving
  generalization, 2021.

\bibitem[French(1999)]{catastrophic_forgetting}
Robert French.
\newblock Catastrophic forgetting in connectionist networks.
\newblock \emph{Trends in cognitive sciences}, 3:\penalty0 128--135, 1999.

\bibitem[Garipov et~al.(2018)Garipov, Izmailov, Podoprikhin, Vetrov, and
  Wilson]{garipov2018loss}
Timur Garipov, Pavel Izmailov, Dmitrii Podoprikhin, Dmitry~P Vetrov, and
  Andrew~G Wilson.
\newblock Loss surfaces, mode connectivity, and fast ensembling of dnns.
\newblock In \emph{Advances in Neural Information Processing Systems}. Curran
  Associates, Inc., 2018.

\bibitem[Gu et~al.(2022)Gu, Yang, Wei, and Deng]{Gu:2022}
Yanan Gu, Xu Yang, Kun Wei, and Cheng Deng.
\newblock Not just selection, but exploration: Online class-incremental
  continual learning via dual view consistency.
\newblock In \emph{{IEEE/CVF} Conference on Computer Vision and Pattern
  Recognition, {CVPR} 2022, New Orleans, LA, USA, June 18-24, 2022}, pages
  7432--7441. {IEEE}, 2022.

\bibitem[He et~al.(2015)He, Zhang, Ren, and Sun]{DBLP:journals/corr/HeZRS15}
Kaiming He, Xiangyu Zhang, Shaoqing Ren, and Jian Sun.
\newblock Deep residual learning for image recognition.
\newblock \emph{CoRR}, abs/1512.03385, 2015.

\bibitem[Hinton et~al.(2015)Hinton, Vinyals, and Dean]{hinton2015distilling}
Geoffrey Hinton, Oriol Vinyals, and Jeff Dean.
\newblock Distilling the knowledge in a neural network, 2015.

\bibitem[Huang et~al.(2021)Huang, Liang, Liang, and He]{Huang:2021}
Zhongzhan Huang, Mingfu Liang, Senwei Liang, and Wei He.
\newblock Altersgd: Finding flat minima for continual learning by alternative
  training.
\newblock \emph{CoRR}, abs/2107.05804, 2021.

\bibitem[Jordan et~al.(2022)Jordan, Sedghi, Saukh, Entezari, and
  Neyshabur]{jordan2022repair}
Keller Jordan, Hanie Sedghi, Olga Saukh, Rahim Entezari, and Behnam Neyshabur.
\newblock Repair: Renormalizing permuted activations for interpolation repair,
  2022.

\bibitem[Karhadkar et~al.(2024)Karhadkar, Murray, Tseran, and
  Montúfar]{karhadkar2024mildly}
Kedar Karhadkar, Michael Murray, Hanna Tseran, and Guido Montúfar.
\newblock Mildly overparameterized relu networks have a favorable loss
  landscape, 2024.

\bibitem[Kirkpatrick et~al.(2016)Kirkpatrick, Pascanu, Rabinowitz, Veness,
  Desjardins, Rusu, Milan, Quan, Ramalho, Grabska{-}Barwinska, Hassabis,
  Clopath, Kumaran, and Hadsell]{DBLP:journals/corr/KirkpatrickPRVD16}
James Kirkpatrick, Razvan Pascanu, Neil~C. Rabinowitz, Joel Veness, Guillaume
  Desjardins, Andrei~A. Rusu, Kieran Milan, John Quan, Tiago Ramalho, Agnieszka
  Grabska{-}Barwinska, Demis Hassabis, Claudia Clopath, Dharshan Kumaran, and
  Raia Hadsell.
\newblock Overcoming catastrophic forgetting in neural networks.
\newblock \emph{CoRR}, abs/1612.00796, 2016.

\bibitem[Korycki and Krawczyk(2021)]{Korycki:2021}
Lukasz Korycki and Bartosz Krawczyk.
\newblock Class-incremental experience replay for continual learning under
  concept drift.
\newblock In \emph{{IEEE} Conference on Computer Vision and Pattern Recognition
  Workshops, {CVPR} Workshops}, pages 3649--3658, 2021.

\bibitem[Krizhevsky(2009)]{Krizhevsky09learningmultiple}
Alex Krizhevsky.
\newblock Learning multiple layers of features from tiny images.
\newblock Technical report, 2009.

\bibitem[Kühnel et~al.(2023)Kühnel, Schulz, Hammer, and
  Fluck]{kuhnel2023bert}
Lisa Kühnel, Alexander Schulz, Barbara Hammer, and Juliane Fluck.
\newblock Bert weaver: Using weight averaging to enable lifelong learning for
  transformer-based models in biomedical semantic search engines, 2023.

\bibitem[Lee et~al.(2017)Lee, Yoon, Yang, and
  Hwang]{DBLP:journals/corr/abs-1708-01547}
Jeongtae Lee, Jaehong Yoon, Eunho Yang, and Sung~Ju Hwang.
\newblock Lifelong learning with dynamically expandable networks.
\newblock \emph{CoRR}, abs/1708.01547, 2017.

\bibitem[Li et~al.(2017)Li, Xu, Taylor, and
  Goldstein]{DBLP:journals/corr/abs-1712-09913}
Hao Li, Zheng Xu, Gavin Taylor, and Tom Goldstein.
\newblock Visualizing the loss landscape of neural nets.
\newblock \emph{CoRR}, abs/1712.09913, 2017.

\bibitem[Li and Hoiem(2016)]{DBLP:journals/corr/LiH16e}
Zhizhong Li and Derek Hoiem.
\newblock Learning without forgetting.
\newblock \emph{CoRR}, abs/1606.09282, 2016.

\bibitem[Lomonaco et~al.(2021)Lomonaco, Pellegrini, Cossu, Carta, Graffieti,
  Hayes, Lange, Masana, Pomponi, van~de Ven, Mundt, She, Cooper, Forest,
  Belouadah, Calderara, Parisi, Cuzzolin, Tolias, Scardapane, Antiga, Amhad,
  Popescu, Kanan, van~de Weijer, Tuytelaars, Bacciu, and
  Maltoni]{DBLP:journals/corr/abs-2104-00405}
Vincenzo Lomonaco, Lorenzo Pellegrini, Andrea Cossu, Antonio Carta, Gabriele
  Graffieti, Tyler~L. Hayes, Matthias~De Lange, Marc Masana, Jary Pomponi,
  Gido~M. van~de Ven, Martin Mundt, Qi She, Keiland~W. Cooper, Jeremy Forest,
  Eden Belouadah, Simone Calderara, German~Ignacio Parisi, Fabio Cuzzolin,
  Andreas~S. Tolias, Simone Scardapane, Luca Antiga, Subutai Amhad, Adrian
  Popescu, Christopher Kanan, Joost van~de Weijer, Tinne Tuytelaars, Davide
  Bacciu, and Davide Maltoni.
\newblock Avalanche: an end-to-end library for continual learning.
\newblock \emph{CoRR}, abs/2104.00405, 2021.

\bibitem[Lopez-Paz and Ranzato(2017)]{DBLP:journals/corr/Lopez-PazR17}
David Lopez-Paz and Marc\textquotesingle~Aurelio Ranzato.
\newblock Gradient episodic memory for continual learning.
\newblock In \emph{Advances in Neural Information Processing Systems}. Curran
  Associates, Inc., 2017.

\bibitem[Marouf et~al.(2024)Marouf, Roy, Tartaglione, and
  Lathuilière]{marouf2024weighted}
Imad~Eddine Marouf, Subhankar Roy, Enzo Tartaglione, and Stéphane
  Lathuilière.
\newblock Weighted ensemble models are strong continual learners, 2024.

\bibitem[Micikevicius et~al.(2017)Micikevicius, Narang, Alben, Diamos, Elsen,
  Garc{\'{\i}}a, Ginsburg, Houston, Kuchaiev, Venkatesh, and
  Wu]{DBLP:journals/corr/abs-1710-03740}
Paulius Micikevicius, Sharan Narang, Jonah Alben, Gregory~F. Diamos, Erich
  Elsen, David Garc{\'{\i}}a, Boris Ginsburg, Michael Houston, Oleksii
  Kuchaiev, Ganesh Venkatesh, and Hao Wu.
\newblock Mixed precision training.
\newblock \emph{CoRR}, abs/1710.03740, 2017.

\bibitem[Mirzadeh et~al.(2020)Mirzadeh, Farajtabar, G{\"{o}}r{\"{u}}r, Pascanu,
  and Ghasemzadeh]{DBLP:journals/corr/abs-2010-04495}
Seyed{-}Iman Mirzadeh, Mehrdad Farajtabar, Dilan G{\"{o}}r{\"{u}}r, Razvan
  Pascanu, and Hassan Ghasemzadeh.
\newblock Linear mode connectivity in multitask and continual learning.
\newblock \emph{CoRR}, abs/2010.04495, 2020.

\bibitem[Mirzadeh et~al.(2022)Mirzadeh, Chaudhry, Yin, Nguyen, Pascanu,
  G{\"{o}}r{\"{u}}r, and Farajtabar]{DBLP:journals/corr/abs-2202-00275}
Seyed{-}Iman Mirzadeh, Arslan Chaudhry, Dong Yin, Timothy Nguyen, Razvan
  Pascanu, Dilan G{\"{o}}r{\"{u}}r, and Mehrdad Farajtabar.
\newblock Architecture matters in continual learning.
\newblock \emph{CoRR}, abs/2202.00275, 2022.

\bibitem[Nagarajan and Kolter(2019)]{DBLP:journals/corr/abs-1902-04742}
Vaishnavh Nagarajan and J.~Zico Kolter.
\newblock Uniform convergence may be unable to explain generalization in deep
  learning.
\newblock In \emph{Advances in Neural Information Processing Systems}. Curran
  Associates, Inc., 2019.

\bibitem[Olpadkar and Gavas(2021)]{Olpadkar:2021}
Kaustubh Olpadkar and Ekta Gavas.
\newblock Center loss regularization for continual learning.
\newblock \emph{CoRR}, abs/2110.11314, 2021.

\bibitem[Park et~al.(2019)Park, Hong, Han, and Lee]{Park:2019}
Dongmin Park, Seokil Hong, Bohyung Han, and Kyoung~Mu Lee.
\newblock Continual learning by asymmetric loss approximation with single-side
  overestimation.
\newblock In \emph{2019 {IEEE/CVF} International Conference on Computer Vision,
  {ICCV} 2019, Seoul, Korea (South), October 27 - November 2, 2019}, pages
  3334--3343. {IEEE}, 2019.

\bibitem[Pascanu et~al.(2014)Pascanu, Dauphin, Ganguli, and
  Bengio]{DBLP:journals/corr/PascanuDGB14}
Razvan Pascanu, Yann~N. Dauphin, Surya Ganguli, and Yoshua Bengio.
\newblock On the saddle point problem for non-convex optimization.
\newblock \emph{CoRR}, abs/1405.4604, 2014.

\bibitem[Peña et~al.(2022)Peña, Medeiros, Dubail, Aminbeidokhti, Granger, and
  Pedersoli]{peña2022rebasin}
Fidel A.~Guerrero Peña, Heitor~Rapela Medeiros, Thomas Dubail, Masih
  Aminbeidokhti, Eric Granger, and Marco Pedersoli.
\newblock Re-basin via implicit sinkhorn differentiation, 2022.

\bibitem[Prabhu et~al.(2020{\natexlab{a}})Prabhu, Torr, and
  Dokania]{prabhu2020gdumb}
Ameya Prabhu, Philip~HS Torr, and Puneet~K Dokania.
\newblock Gdumb: A simple approach that questions our progress in continual
  learning.
\newblock In \emph{Computer Vision--ECCV 2020: 16th European Conference,
  Glasgow, UK, August 23--28, 2020, Proceedings, Part II 16}, pages 524--540.
  Springer, 2020{\natexlab{a}}.

\bibitem[Prabhu et~al.(2020{\natexlab{b}})Prabhu, Torr, and
  Dokania]{10.1007/978-3-030-58536-5_31}
Ameya Prabhu, Philip H.~S. Torr, and Puneet~K. Dokania.
\newblock Gdumb: A simple approach that questions our progress in continual
  learning.
\newblock In \emph{Computer Vision -- ECCV 2020}, pages 524--540, Cham,
  2020{\natexlab{b}}. Springer International Publishing.

\bibitem[Rebuffi et~al.(2016)Rebuffi, Kolesnikov, and
  Lampert]{DBLP:journals/corr/RebuffiKL16}
Sylvestre{-}Alvise Rebuffi, Alexander Kolesnikov, and Christoph~H. Lampert.
\newblock icarl: Incremental classifier and representation learning.
\newblock \emph{CoRR}, abs/1611.07725, 2016.

\bibitem[Rusu et~al.(2016)Rusu, Rabinowitz, Desjardins, Soyer, Kirkpatrick,
  Kavukcuoglu, Pascanu, and Hadsell]{DBLP:journals/corr/RusuRDSKKPH16}
Andrei~A. Rusu, Neil~C. Rabinowitz, Guillaume Desjardins, Hubert Soyer, James
  Kirkpatrick, Koray Kavukcuoglu, Razvan Pascanu, and Raia Hadsell.
\newblock Progressive neural networks.
\newblock \emph{CoRR}, abs/1606.04671, 2016.

\bibitem[Sagun et~al.(2016)Sagun, Bottou, and
  LeCun]{DBLP:journals/corr/SagunBL16}
Levent Sagun, L{\'{e}}on Bottou, and Yann LeCun.
\newblock Singularity of the hessian in deep learning.
\newblock \emph{CoRR}, abs/1611.07476, 2016.

\bibitem[Sandler et~al.(2018)Sandler, Howard, Zhu, Zhmoginov, and
  Chen]{DBLP:journals/corr/abs-1801-04381}
Mark Sandler, Andrew~G. Howard, Menglong Zhu, Andrey Zhmoginov, and
  Liang{-}Chieh Chen.
\newblock Inverted residuals and linear bottlenecks: Mobile networks for
  classification, detection and segmentation.
\newblock \emph{CoRR}, abs/1801.04381, 2018.

\bibitem[Sankar et~al.(2020)Sankar, Khasbage, Vigneswaran, and
  Balasubramanian]{Sankar2020ADL}
Adepu~Ravi Sankar, Yash Khasbage, Rahul Vigneswaran, and Vineeth~N.
  Balasubramanian.
\newblock A deeper look at the hessian eigenspectrum of deep neural networks
  and its applications to regularization.
\newblock \emph{ArXiv}, abs/2012.03801, 2020.

\bibitem[Schwarz et~al.(2018)Schwarz, Luketina, Czarnecki, Grabska-Barwinska,
  Teh, Pascanu, and Hadsell]{schwarz2018progress}
Jonathan Schwarz, Jelena Luketina, Wojciech~M. Czarnecki, Agnieszka
  Grabska-Barwinska, Yee~Whye Teh, Razvan Pascanu, and Raia Hadsell.
\newblock Progress \& compress: A scalable framework for continual learning,
  2018.

\bibitem[Stoica et~al.(2024)Stoica, Bolya, Bjorner, Ramesh, Hearn, and
  Hoffman]{stoica2024zipit}
George Stoica, Daniel Bolya, Jakob Bjorner, Pratik Ramesh, Taylor Hearn, and
  Judy Hoffman.
\newblock Zipit! merging models from different tasks without training, 2024.

\bibitem[Tan and Le(2019)]{DBLP:journals/corr/abs-1905-11946}
Mingxing Tan and Quoc~V. Le.
\newblock Efficientnet: Rethinking model scaling for convolutional neural
  networks.
\newblock \emph{CoRR}, abs/1905.11946, 2019.

\bibitem[Tatro et~al.(2020)Tatro, Chen, Das, Melnyk, Sattigeri, and
  Lai]{DBLP:journals/corr/abs-2009-02439}
N.~Joseph Tatro, Pin{-}Yu Chen, Payel Das, Igor Melnyk, Prasanna Sattigeri, and
  Rongjie Lai.
\newblock Optimizing mode connectivity via neuron alignment.
\newblock \emph{CoRR}, abs/2009.02439, 2020.

\bibitem[van~de Ven and Tolias(2019)]{DBLP:journals/corr/abs-1904-07734}
Gido~M. van~de Ven and Andreas~S. Tolias.
\newblock Three scenarios for continual learning.
\newblock \emph{CoRR}, abs/1904.07734, 2019.

\bibitem[Wu(2017)]{Wu2017TinyIC}
Jiayu Wu.
\newblock Tiny imagenet challenge.
\newblock 2017.

\bibitem[Wu et~al.(2019)Wu, Chen, Wang, Ye, Liu, Guo, and Fu]{Wu2019LargeSI}
Yue Wu, Yinpeng Chen, Lijuan Wang, Yuancheng Ye, Zicheng Liu, Yandong Guo, and
  Yun~Raymond Fu.
\newblock Large scale incremental learning.
\newblock \emph{2019 IEEE/CVF Conference on Computer Vision and Pattern
  Recognition (CVPR)}, pages 374--382, 2019.

\bibitem[Yan et~al.(2021)Yan, Xie, and He]{DBLP:journals/corr/abs-2103-16788}
Shipeng Yan, Jiangwei Xie, and Xuming He.
\newblock {DER:} dynamically expandable representation for class incremental
  learning.
\newblock \emph{CoRR}, abs/2103.16788, 2021.

\bibitem[Zagoruyko and Komodakis(2016)]{DBLP:journals/corr/ZagoruykoK16}
Sergey Zagoruyko and Nikos Komodakis.
\newblock Wide residual networks.
\newblock \emph{CoRR}, abs/1605.07146, 2016.

\bibitem[Zenke et~al.(2017{\natexlab{a}})Zenke, Poole, and
  Ganguli]{DBLP:journals/corr/ZenkePG17}
Friedemann Zenke, Ben Poole, and Surya Ganguli.
\newblock Improved multitask learning through synaptic intelligence.
\newblock \emph{CoRR}, abs/1703.04200, 2017{\natexlab{a}}.

\bibitem[Zenke et~al.(2017{\natexlab{b}})Zenke, Poole, and
  Ganguli]{pmlr-v70-zenke17a}
Friedemann Zenke, Ben Poole, and Surya Ganguli.
\newblock Continual learning through synaptic intelligence.
\newblock In \emph{Proceedings of the 34th International Conference on Machine
  Learning}, pages 3987--3995. PMLR, 2017{\natexlab{b}}.

\end{thebibliography}
}


\clearpage
\setcounter{page}{1}
\maketitlesupplementary

\section{CLeWI without rehearsal}
\label{sec:clewi_without_rehearsal}

\begin{figure*}
    \centering
    \includegraphics[width=\linewidth]{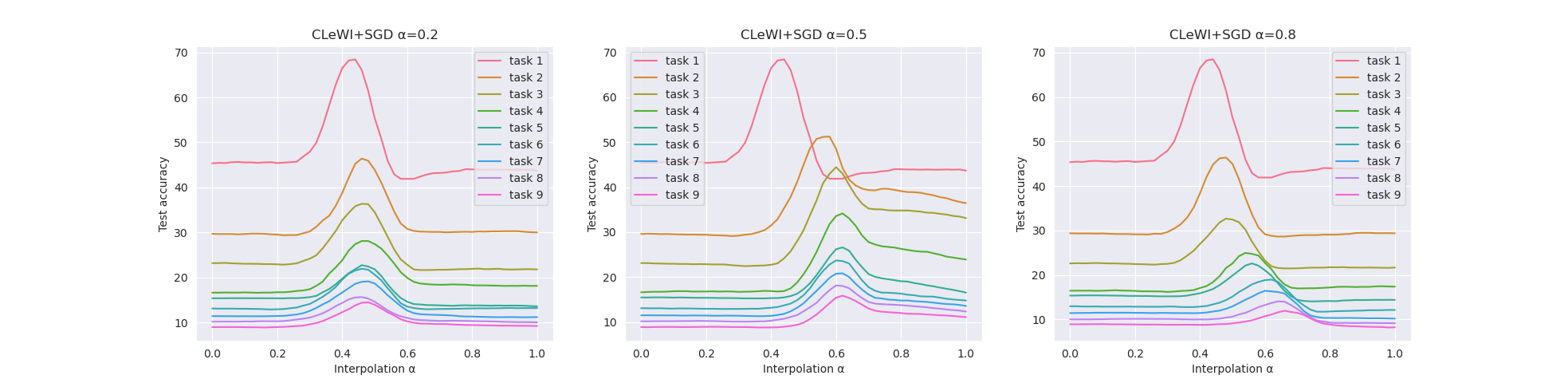}
    \caption{Test accuracy in the function of interpolation alpha for CLeWI with no replay}
    \label{fig:interpolation_no_rehersal}
\end{figure*}

\begin{table}[]
    \centering
    \caption{Average accuracy and forgetting measure for CLeWI without rehearsal.}
    \begin{tabular}{c|c|c|c}
        method & Acc($\uparrow$) & Acc$_K$ ($\uparrow$) & FM ($\downarrow$)  \\
        \hline
 CLeWI-SGD $\alpha=0.2$ & 8.95  & 88.3 & 80.66 \\
 CLeWI-SGD $\alpha=0.5$ & 9.87  & 85.7 & 71.04 \\
 CLeWI-SGD $\alpha=0.6$ & 11.66 & 81.1 & 54.08 \\
 CLeWI-SGD $\alpha=0.7$ & 16.41 & 62.2 & 22.0  \\
 CLeWI-SGD $\alpha=0.8$ & 8.87  & 3.1  & 1.04  \\
        \hline
    \end{tabular}
    \label{tab:no_rehersal}
\end{table}

In preliminary experiments, we started by applying the weight interpolation without any other form of forgetting prevention. In such an approach, we kept the buffer in our algorithm, but it was used only to compute activations and batch normalization statistics required by REPAIR. Data from the buffer was not used for network training. The results are provided in \cref{tab:no_rehersal}. We tried several settings of $\alpha$, but none provided significant gains in accuracy. With $\alpha = 0.2$ or $\alpha = 0.5$, the model forgets completely what it has learned previously, as illustrated by the high accuracy on the last task and low forgetting. By using $\alpha=0.8$, we could almost entirely eliminate catastrophic forgetting, but only for the first task. The model lacks the plasticity to learn new tasks. The best results are obtained for $\alpha=0.7$. However, they are far worse than any other method that stores previous data in a buffer.

We also provide the interpolation plots of CLeWI-SGD for the several values of $\alpha$ in \cref{fig:interpolation_no_rehersal}. The local maxima of interpolation plots move significantly when different values of $\alpha$ are used. This was not observed for other continual learning algorithms that CLeWI was combined with. We can also see here that the model with $\alpha=0.6$ or $0.7$ should obtain higher accuracy, as this value aligns well with the local maxima of the plot. We obtain the best results as shown by \cref{tab:no_rehersal} for these $\alpha$ values. 

\section{Interpolation plots for other algorithms}
\label{sec:interpolation_others}

\begin{figure*}
    \centering
    \includegraphics[width=1.0\linewidth]{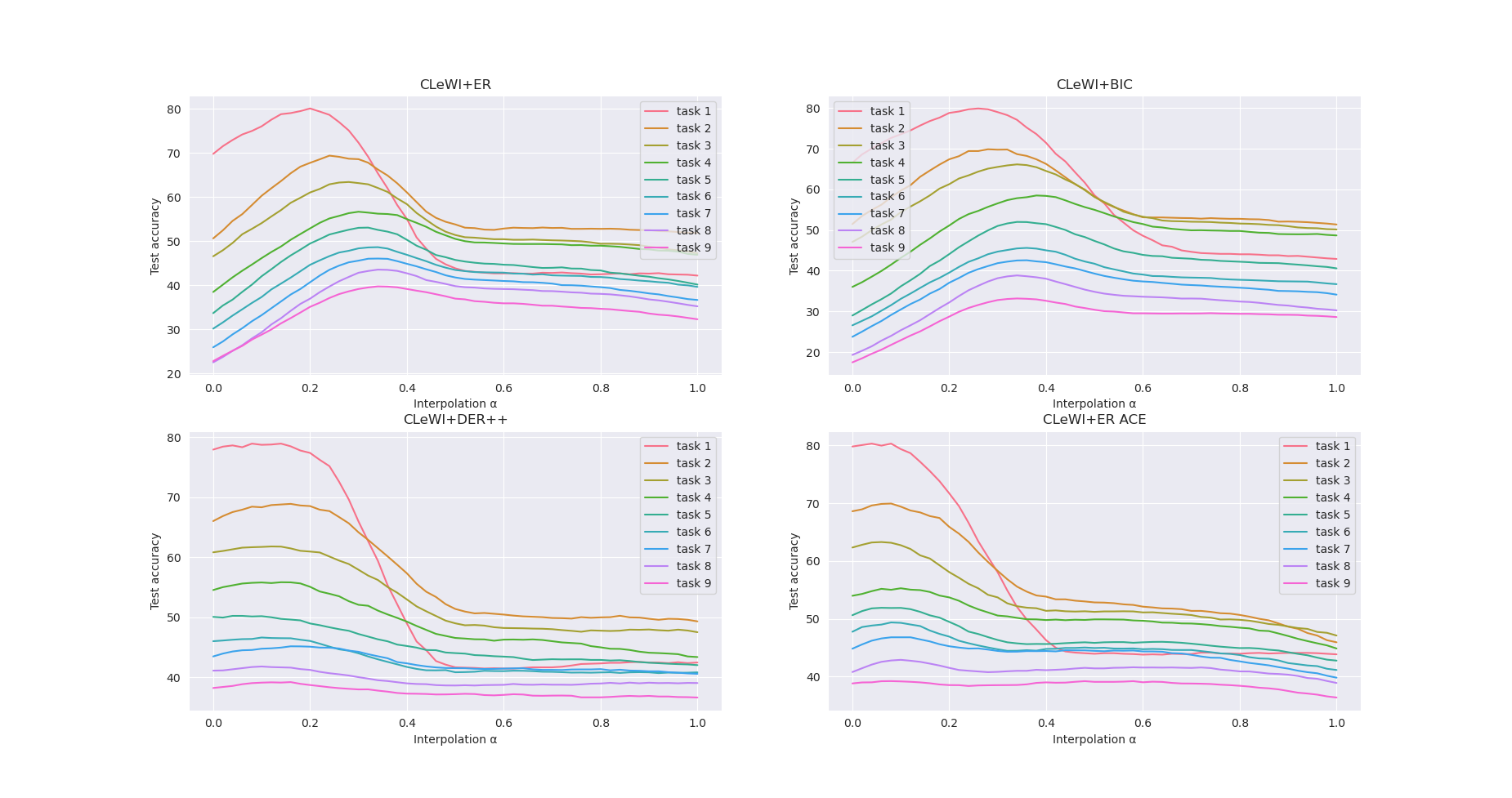}
    \caption{Interpolation plots for other forms of rehearsal}
    \label{fig:others_interpolation}
\end{figure*}

We present interpolation plots for various forms of rehearsal in \cref{fig:others_interpolation}. These plots show that each form of rehearsal requires a different interpolation $\alpha$ hyperparameter. These could be obtained with either a hyperparameter search, and interpolation plots can be helpful in this process. In experiments with standard benchmarks, we have tuned $\alpha$ only once and used it across different datasets.

We provide the exact values of the interpolation $\alpha$ in \cref{tab:alpha_values} for completeness.

\begin{table}[]
    \centering
    \caption{Values of interpolation $\alpha$ hyperparameter used in experiments}
    \begin{tabular}{c|c}
        method & $\alpha$ \\
        \hline
        CLeWI ER & 0.3 \\
        CLeWI aGEM & 0.5 \\
        CLeWI ER ACE & 0.3 \\
        CLeWI MIR & 0.5 \\
        CLeWI BIC & 0.5 \\
        CLeWI DER++ & 0.2 \\
        \hline
    \end{tabular}
    \label{tab:alpha_values}
\end{table}


\section{Interpolation and REPAIR algorithm details}
\label{sec:repair}

In this section, we provide details about the interpolation process and REPAIR \cite{jordan2022repair} algorithm used in our experiments. Both of these steps are represented in pseudocode as functions \emph{calc\_permutation } and \emph{update\_batchnorm}. In the first step, the optimal permutation is found. In the second step, batch normalization is performed in order to mitigate variance collapse.
\\
\textbf{Step 1 - alignment} In the first step, we search for the permutation $\pi$ of $\theta_P$ that maximizes the correlation between feature maps from $\theta_P$ and $\theta$ networks. The feature maps are obtained from samples in buffer $\mathcal{M}$. Note that $\mathcal{M}$ contains information about all tasks. Specifically, for a given layer, having the feature maps of the dimension $N \times C \times W \times H$ from two networks, we first calculate the $C \times C$ correlation matrix between them. Next, we choose the optimal permutation by solving the linear sum assignment problem with an optimizer from \textit{scipy}. In our experiments to obtain the activations, we use only a single epoch with a batch size equal to~32. 
\\
\textbf{Step 2 - normalization} Aligned networks suffer from the phenomenon called variance collapse \cite{jordan2022repair}. The variance of the feature maps is decreasing with network depth leading to poor performance. We prevent variance collapse by renormalizing feature maps, ensuring the variances of feature maps in the interpolated network $\theta_\alpha$ satisfy conditions: $EX_\alpha = (1-\alpha) EX + \alpha EX_P$ and $\mathrm{Var}{X_\alpha} = (1-\alpha) \mathrm{Var}{X} + \alpha \mathrm{Var}{X_P}$, where $X_\alpha$, $X_P$, $X$ are random variables corresponding to feature maps of $\theta_\alpha$ (interpolated), $\theta_P$ (trained on previous tasks and permuted), and $\theta$ (trained on current task) networks respectively. To this aim, after all interpolated layers, Batch Normalization layers are added to apply an affine transformation to feature maps of the network $\theta_\alpha$. Parameters of the affine transformation are set to the interpolations of means and variances of feature maps in networks $\theta_P$ and $\theta$, producing feature maps with means and standard deviations satisfying conditions to prevent the variance collapse. After that, additional Batch Normalization layers are removed from the network via BatchNorm fusion. In our paper we use the observation from \cite{jordan2022repair}, that in architectures with Batch Normalization after each layer fixing the variance collapse via additional Batch Normalizations is equivalent to resetting the \textit{batch\_norm} statistics. We use a single epoch with data from $\mathcal{M}$ buffer to perform this reset.

\section{Memory restricted evaluation of rehearsal methods}
\label{sec:memory_restricted}

Some researchers in the continual learning community argue that a more fair comparison between algorithms would be assigning the same amount of memory for the buffer for all algorithms. All objects stored in memory, such as images, models, activations, or weights, would use this memory until the buffer is full.  We are aware that our model requires additional memory for weights. For this reason, we carry out additional experiments in this memory-restricted evaluation mode to check if storing additional weights can bring improvement over increasing the buffer size alone.

\begin{table*}[!tbp]
    \centering
    \caption{Buffer size and accuracy for various continual learning algorithms and architectures for memory-restricted evaluation of CLeWI.}
    \begin{tabular}{c|c|c|c|c|c}
        \hline
        \multirow{2}{*}{method} & \multirow{2}{*}{backbone} & \multicolumn{2}{c}{Cifar100(T=10)} & \multicolumn{2}{c}{Tiny-ImageNet(T=20)} \\
        \cline{3-6} 
         &  & \makecell{\#imgs in\\buffer} & Acc ($\uparrow$) &  \makecell{\#imgs in\\buffer} & Acc ($\uparrow$) \\
        \hline
        ER & \multirow{2}{*}{ResNet18}  & 15000 & 62.61  & 4000 & 21.18 \\
        CLeWI+ER & & 394 & 38.03  & 348 & 9.45 \\
        \hline
        ER & \multirow{2}{*}{MobileNetv2} & 3500 & 26.55  & 1000 & 9.63 \\ 
        CLeWI+ER &  & 438 & 18.67 & 235 & 6.96 \\
        \hline
        \hline
        ER & \multirow{2}{*}{ResNet18} & 15606 & 61.53 & 4652 & 20.84 \\
        CLeWI+ER & & 1000 & 46.72 & 1000 & 21.52 \\
        \hline
        ER & \multirow{2}{*}{MobileNetv2} & 4062 & 27.37 & 1765 & 7.42 \\ 
        CLeWI+ER &  & 1000 & 24.72 & 1000 & 12.16 \\
        \hline
    \end{tabular}
    \label{tab:memory_restricted}
\end{table*}

\begin{figure}
    \centering
    \includegraphics[width=1.0\linewidth]{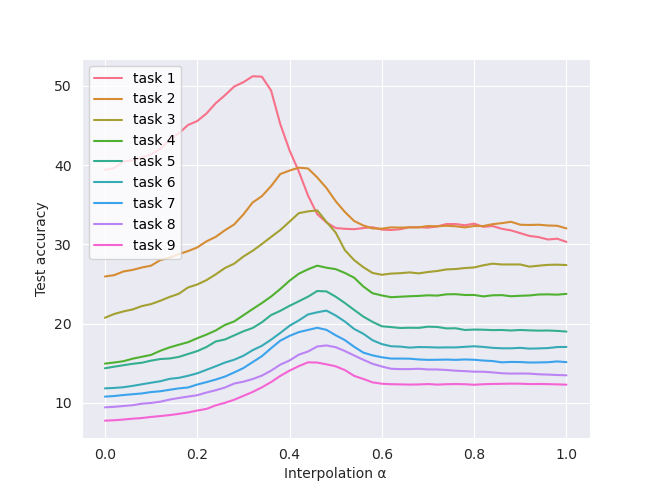}
    \caption{Test accuracy in the function of interpolation $\alpha$ for MobileNetv2 architecture}
    \label{fig:mobilenet}
\end{figure}

As thorough evaluation requires several architectures, we employ both reduced ResNet architecture \cite{DBLP:journals/corr/HeZRS15} with an overall number of parameters equal to 11220132 and MobileNetv2 \cite{DBLP:journals/corr/abs-1801-04381} architecture with 2351972 parameters.
To enable interpolation for MobileNetv2, we made a custom implementation of the REPAIR algorithm for this architecture. We are not applying the permutation for the depthwise convolution, as the same filter is applied to all input channels. For all other layers, permutation is applied. To verify that the algorithm works for other architecture, we made an interpolation plot, presented in \cref{fig:mobilenet}. This figure shows that interpolation can still improve test set accuracy during training with the MobileNetv2 backbone.

We treat each network parameter as 32bit variable. We do not use mixed precision training \cite{DBLP:journals/corr/abs-1710-03740}, or other methods of reducing weights memory footprint. For Cifar100, each image is an array of size 32x32x3, with each color saved as an 8-bit variable. Analogically, we treat each Tiny-ImageNet image as a 64x64x3 array with 8-bit pixels.

From these assumptions, we can calculate the memory-wise equivalent of storing model weights in terms of additional images, which could be used in the rehearsal algorithm. 
For example, we computed that saving one ResNet corresponds to storing 14609 additional CIFAR100 images. 
We compare ER with CLeWI-ER to eliminate the influence of more advanced replay methods from our results. We carry out two experiments. In first we calculate the number of images that could be stored in the memory used by weights, and then we round this number up. CLeWI in this experiment has only the number of images that was rounded up, while ER has the full buffer size, which has the same memory footprint as both the memory buffer of CLeWI and model weights.
In the second mode we increase the number of images stored in the CLeWI buffer to 1000, and increase the ER buffer size by the same amount. 
The results are provided in \cref{tab:memory_restricted}.

Two factors could greatly impact this evaluation mode: the size of the images in the dataset and the number of parameters in the model. For datasets with smaller images, such as CIFAR100, storing 32-bit weights in memory corresponds to a huge increase in the buffer size. In such cases, matching the ER's performance with a big buffer is hard.
When we consider more parameter-efficient models, such as MobielNetv2, the gap between ER and CLeWI becomes smaller. The difference in the buffer size is also smaller, and the accuracy of CLeWI, especially for Tiny-ImageNet, becomes closer to ER.
When we allow an increase in the buffer size for both algorithms, the gap between ER and CLeWI becomes even smaller, and we even notice accuracy improvement for the Tiny-ImageNet dataset. An increase in buffer size does not significantly impact the ER, as performance there could already be saturated.

Please note that here, we carry out experiments with benchmarks for continual learning used in the previous parts of the paper. If we consider other datasets with higher image sizes, such as 224x224 or 512×512, the comparison could be even more favorable for CLeWI. 
We may also employ more recent architectures for training, such as EfficientNetB0 \cite{DBLP:journals/corr/abs-1905-11946}, and utilize mixed precision training \cite{DBLP:journals/corr/abs-1710-03740} to further reduce the memory footprint of the backbone.

We conclude that in a memory-restricted mode of evaluation, CLeWI could improve over vanilla ER, but only if the images are big and the network architecture is parameter efficient. In other cases, it could be more advisable to increase the buffer size alone and not use interpolation.

\section{Implementation}
\label{sec:implementation}

The CLeWI algorithm is straightforward to implement and can be used as a plug-in for existing continual learning algorithms. This is in line with the construction of modern continual learning frameworks such as Mammoth \cite{buzzega2020dark}, or Avalanche \cite{DBLP:journals/corr/abs-2104-00405}. In such a case, one needs to implement only a method that is called after training with a given task that performs interpolation and stores the previous model. This should allow for easy combining of existing methods for improved performance. However, one needs to be careful during implementation. Other algorithms could also take some actions after training with a given task. Whether these actions are taken before or after the interpolation could impact the overall algorithm performance. This should be considered case-by-case, as different algorithms could require different orders of post-task training actions. In many cases, this can be trivial, as other algorithms do not collide with weight interpolation, but there are a few that require extra steps in the implementation. For example, we found that in the case of BiC 
\cite{Wu2019LargeSI}, first, we need to compute the bias correction, then store the old model, then update the buffer, perform weight interpolation, and evaluate bias compute bias correction for interpolated models again. As a result, for BiC+CLeWI, we keep three copies of models instead of two: one for the current model, the second for BiC knowledge distillation, and the last for weight interpolation. Using the same model for weight interpolation and knowledge distillation provided worse results. The model before interpolation has better representation of most recent task, therefore could be more suitable for knowledge destination. This should be taken into consideration when combining CLeWI with other algorithms that store weights our utilize activations of a network trained on previous tasks (such as LwF \cite{DBLP:journals/corr/LiH16e}). Combining CLeWI with other methods that require post-training actions may require similar modifications.










\end{document}